\pdfoutput=1
\documentclass[letterpaper]{article} 
\usepackage{aaai2026}  
\usepackage{times}  
\usepackage{helvet}  
\usepackage{courier}  
\usepackage[hyphens]{url}  
\usepackage{graphicx} 
\usepackage{natbib}  
\usepackage{caption} 
\usepackage{algorithm}
\usepackage{algorithmic}

\usepackage{bibunits}
\usepackage[most]{tcolorbox}

\usepackage{newfloat}
\usepackage{listings}
\DeclareCaptionStyle{ruled}{labelfont=normalfont,labelsep=colon,strut=off} 
\floatstyle{ruled}
\newfloat{listing}{tb}{lst}{}
\floatname{listing}{Listing}

\usepackage{xcolor}
\usepackage[most]{tcolorbox}
\usepackage{graphicx}
\usepackage{caption}
\usepackage{booktabs}
\usepackage{multirow}
\usepackage{arydshln} 

\definecolor{searchcolor}{RGB}{0,102,204}
\definecolor{thinkcolor}{RGB}{153,0,153}
\definecolor{infocolor}{RGB}{0,153,51}
\definecolor{answercolor}{RGB}{204,102,0}

\title{REX-RAG: Reasoning Exploration with Policy Correction in Retrieval-Augmented Generation}
\author {
    Wentao Jiang\textsuperscript{\rm 1}\thanks{Equal Contribution},
    Xiang Feng\textsuperscript{\rm 1\rm *},
    Zengmao Wang\textsuperscript{\rm 1\rm †},
    Yong Luo\textsuperscript{\rm 1},
    Pingbo Xu\textsuperscript{\rm 2, \rm 3},
    Zhe Chen\textsuperscript{\rm 4},
    Bo Du\textsuperscript{\rm 1},\\
    Jing Zhang\textsuperscript{\rm 1}\thanks{Corresponding Author}
}
\affiliations {
    \textsuperscript{\rm 1}School of Computer Science, Wuhan University, China\\
    \textsuperscript{\rm 2}Department of Anesthesiology, Zhejiang Cancer Hospital, China\\
    \textsuperscript{\rm 3}Institute of Medicine, Chinese Academy of Sciences, Hangzhou, Zhejiang, China\\
    \textsuperscript{\rm 4}Department of Computer Science and Information Technology, La Trobe University, Australia\\
    jiang\_wentao@whu.edu.cn, fengxiang\_cs@whu.edu.cn, wangzengmao@whu.edu.cn, luoyong@whu.edu.cn,  xupingboshanghai@163.com, Zhe.Chen@latrobe.edu.au, dubo@whu.edu.cn, jingzhang.cv@gmail.com
}

\usepackage{amsmath}
\usepackage{amssymb}
\usepackage{amsfonts}
\usepackage{graphicx}

\begin{document}

\maketitle

\begin{abstract}
Reinforcement learning (RL) is emerging as a powerful paradigm for enabling large language models (LLMs) to perform complex reasoning tasks. Recent advances indicate that integrating RL with retrieval-augmented generation (RAG) allows LLMs to dynamically incorporate external knowledge, leading to more informed and robust decision making. However, we identify a critical challenge
during policy-driven trajectory sampling: LLMs are frequently trapped in unproductive reasoning paths, which we refer to as ``dead ends", committing to overconfident yet incorrect conclusions. This severely hampers exploration and undermines effective policy optimization.
To address this challenge, we propose \textbf{REX-RAG} (Reasoning Exploration with Policy Correction in Retrieval-Augmented Generation), a novel framework that explores alternative reasoning paths while maintaining rigorous policy learning through principled distributional corrections. Our approach introduces two key innovations: (1) Mixed Sampling Strategy, which combines a novel probe sampling method with exploratory prompts to escape dead ends; and (2) Policy Correction Mechanism, which employs importance sampling to correct distribution shifts induced by mixed sampling, thereby mitigating gradient estimation bias. We evaluate it on seven question-answering benchmarks, and the experimental results show that REX-RAG achieves average performance gains of 5.1\% on Qwen2.5-3B and 3.6\% on Qwen2.5-7B over strong baselines, demonstrating competitive results across multiple datasets. The code is publicly available at \url{https://github.com/MiliLab/REX-RAG}.

\end{abstract}
\section{Introduction}

Recent advances have shown that reinforcement learning (RL) offers a promising avenue for training large language models (LLMs) to perform complex reasoning tasks~\cite{RLHF, DAPO, ReasoningEra}. By integrating multi-step reasoning with retrieval-augmented generation (RAG), RL-trained LLMs can dynamically leverage external knowledge sources—essentially allowing them to ``think while searching"~\cite{ReSearch, Search-r1}. This paradigm holds particular promise for multi-hop question answering, where models must iteratively gather and synthesize evidence across multiple queries to arrive at well-founded conclusions~\cite{Search-r1-emperical}.

\begin{figure}[t]
  \centering
  \includegraphics[width=0.45\textwidth]{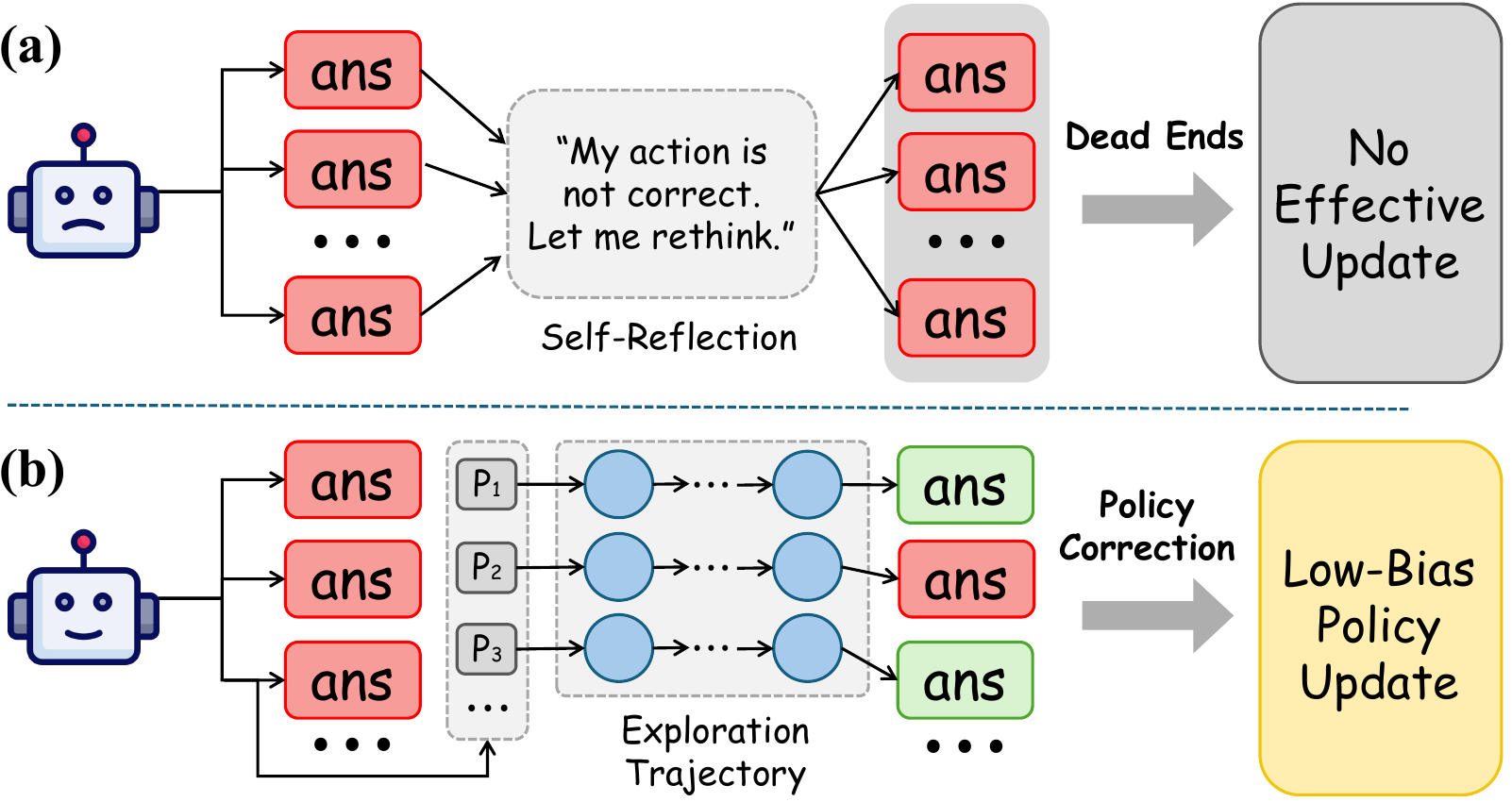}
  \caption{Framework comparison between existing approaches and REX-RAG. (a) Self-Reflection: when encountering incorrect answers, the model attempts to ``rethink", but often produces similar trajectories that lead to dead ends with no effective updates. (b) REX-RAG: our method employs mixed sampling with exploration trajectories guided by diverse reasoning prompts, followed by policy correction to ensure low-bias policy updates.}
  \label{fig:framework_comparison}
\end{figure}

Despite this potential, we observe a critical challenge that substantially hinders policy optimization in such settings. During RL training, LLMs frequently become trapped in what we term ``\textit{dead ends}": situations where the model consistently fails to arrive at the correct final answer after multiple rollouts. This phenomenon often stems from premature or overconfident conclusions drawn despite insufficient supporting information, effectively terminating exploration along potentially fruitful reasoning ~\cite{yue2025does, wen2025reinforcement, liu2025understanding}. 

\begin{figure}[t]
  \centering
  \includegraphics[width=0.4\textwidth]{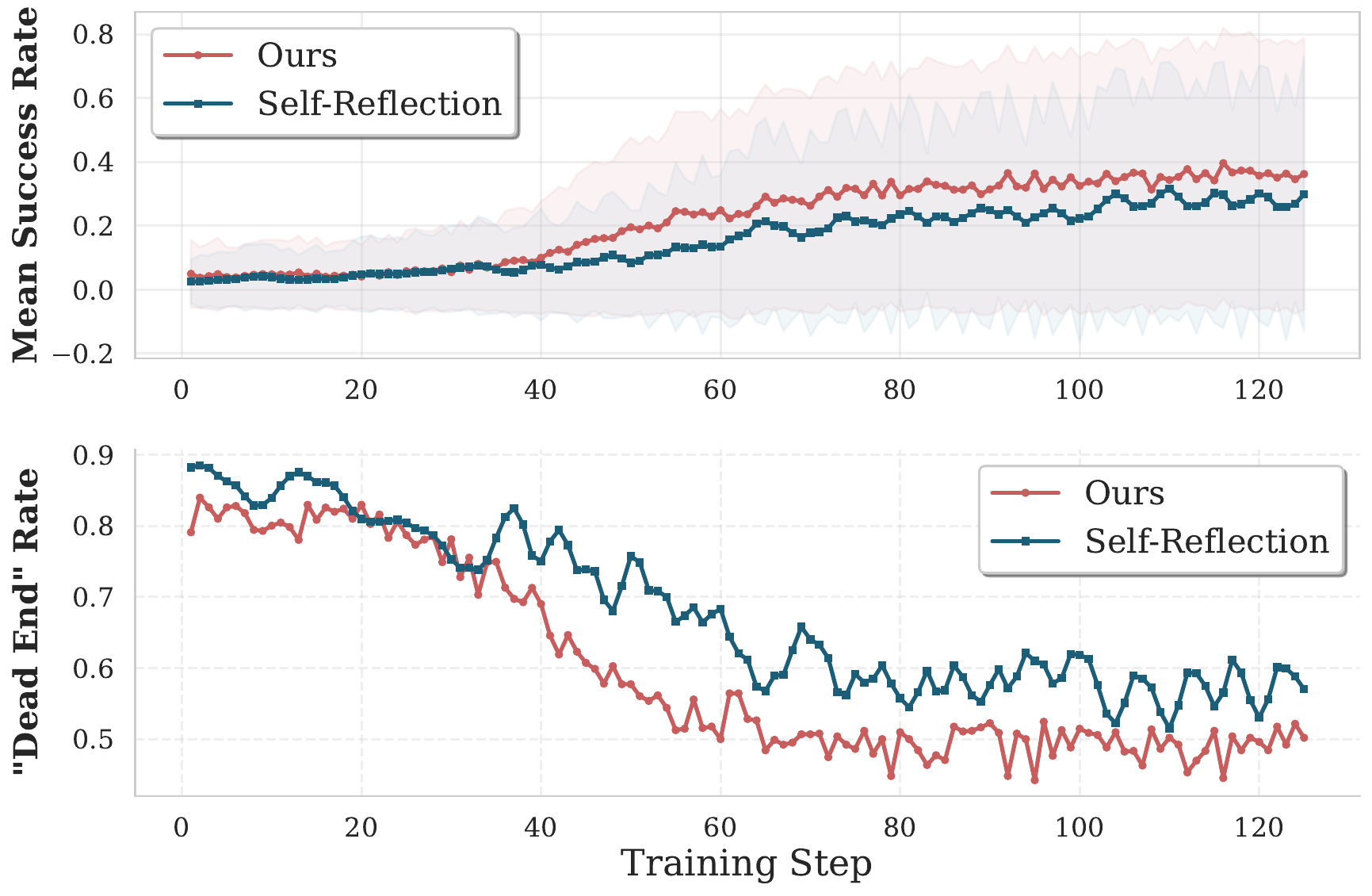}
  \caption{Training dynamics comparison between self-reflection and REX-RAG. Top: Success rate over training steps, showing REX-RAG (red) achieving higher and more stable performance compared to self-reflection (blue). Bottom: Dead end rate over training steps, demonstrating that REX-RAG effectively reduces dead ends throughout while self-reflection shows persistent high ``dead end" rates.}
  \label{fig:training_dynamics}
\end{figure}

Addressing this challenge requires mechanisms that can proactively explore alternative reasoning paths when initial trajectories prove unproductive. A straightforward solution is \textit{self-reflection}~\cite{Deepseek-r1, Search-r1}, which attempts to revise failed reasoning chains to generate alternative ones. However, we observe that these revised trajectories are often merely slight perturbations of the original paths, offering limited novelty and insufficient deviation to meaningfully explore alternative solutions. Consequently, it struggles to escapee from dead-end reasoning paths, as illustrated in Fig.~\ref{fig:framework_comparison}(a). In our experiments with the Qwen2.5-3B model, self-reflection consistently results in a high incidence of ``dead ends", where LLMs generate wrong answers across all rollouts. This phenomenon surpasses 85\% in the early phases of RL training and significantly impedes effective policy learning, as shown in Fig.~\ref{fig:training_dynamics}.

On the other hand, more aggressively enforcing exploration, such as introducing additional agents~\cite{RAG-Gym, MA-RAG}, makes end-to-end optimization challenging due to the complexity of jointly training multiple components. This challenge underscores the need for principled strategies that can foster sufficiently diverse and informative exploration while ensuring stable and unbiased policy optimization without compromising the end-to-end learning paradigm~\cite{GiGPO}.

To address this challenge, we propose \textbf{REX-RAG} (\textbf{R}easoning \textbf{EX}ploration with Policy Correction in Retrieval-Augmented Generation), a novel framework that explores alternative reasoning paths while maintaining rigorous policy learning through principled distributional corrections. Our framework incorporates an exploratory probe policy that collaborates with the standard policy to escape from the ``dead ends", as shown in Fig.~\ref{fig:framework_comparison} (b).

The key innovation of REX-RAG lies in its \textit{Mixed Sampling Strategy} that combines exploration and exploitation in a principled manner. Our framework employs a curated collection of chain-of-thought prompts to inject diverse reasoning directions when trajectories fail. Specifically, when the policy encounters a dead end—indicated by incorrect answers—we strategically insert concise reasoning hints from the prompt pool and resume generation from that point, effectively steering the model toward unexplored solution paths. This approach generates substantially different reasoning trajectories that can escape local optima while maintaining computational efficiency.

Crucially, to prevent the distributional shifts inherent in such interventions from destabilizing training, REX-RAG incorporates a \textit{Policy Correction Mechanism} based on importance sampling theory. This mechanism accurately estimates the likelihood of probe-induced trajectories and applies appropriate corrections to minimize the bias in the policy gradient, under mixed sampling from both the original policy and the probe policy~\cite{LUFFY, ULTRA}.

Extensive experiments on multi-hop question answering benchmarks demonstrate that REX-RAG significantly outperforms existing methods, achieving substantial improvements in both answer accuracy and reasoning quality.  On average, it outperforms strong baselines by 5.1\% on Qwen2.5-3B and 3.6\% on Qwen2.5-7B. Furthermore, as shown in Fig.~\ref{fig:training_dynamics}, our analysis reveals that the framework successfully escapes dead ends while maintaining stable policy learning, with consistently higher success rates and lower dead end rates compared to self-reflection approaches, validating the effectiveness of our principled exploration strategy.

The main contribution can be concluded that:
\begin{itemize}
    \item We identify and formalize the \textit{dead end} problem in RL-based RAG training, demonstrating its significant impact on policy optimization and showing that it affects over 85\% of training instances in early phases.
    \item We propose \textbf{REX-RAG}, a novel framework combining Mixed Sampling Strategies with Policy Correction Mechanism for effective exploration and stable training.
    \item We achieve substantial improvements over strong baselines (5.1\% on Qwen2.5-3B and 3.6\% on Qwen2.5-7B) on multi-hop question answering benchmarks.
\end{itemize}

\section{Related Work}

\paragraph{Retrieval-Augmented Generation.} 
RAG~\cite{RAG} has fundamentally transformed how language models access and utilize external knowledge. The RAG framework combines search engines with generative models, enabling LLMs to ground their responses in retrieved documents~\cite{arslan2024survey}. This paradigm has proven particularly effective for knowledge-intensive tasks where parametric knowledge alone is insufficient~\cite{PopQA}. 
For multi-hop reasoning tasks, several specialized approaches have emerged~\cite{Self-Reflection, gao2025synergizing}, for example,  IRCoT~\cite{IRCOT} interleaves retrieval with chain-of-thought reasoning, allowing models to iteratively gather evidence across multiple reasoning steps. 
However, these methods rely on supervised fine-tuning or simple prompting, limiting their capacity to learn optimal retrieval and reasoning through interaction.

\paragraph{Reinforcement Learning with Verifiable Rewards (RLVR).} 
RLVR has emerged as a popular approach for improving LLM reasoning.
The integration of RL and RAG has opened new avenues for training LLMs to perform complex reasoning tasks~\cite{Deepresearcher, mei2025O2, qian2025scent}. Recent advances include reasoning-oriented models that employ RL to improve step-by-step reasoning capabilities~\cite{Zerosearch, MMSearch-R1, R3-RAG}. In the context of RAG, Search-R1~\cite{Search-r1} represents a pioneering and excellent effort to apply RL for training LLMs to dynamically interact with search engines. However, as noted in empirical studies~\cite{Search-r1-emperical}, existing RL approaches~\cite{R1-Searcher} for reasoning-search interleaved agents face challenges in exploration efficiency and training stability.

\begin{figure*}[t]
  \centering
  \includegraphics[width=0.96\textwidth]{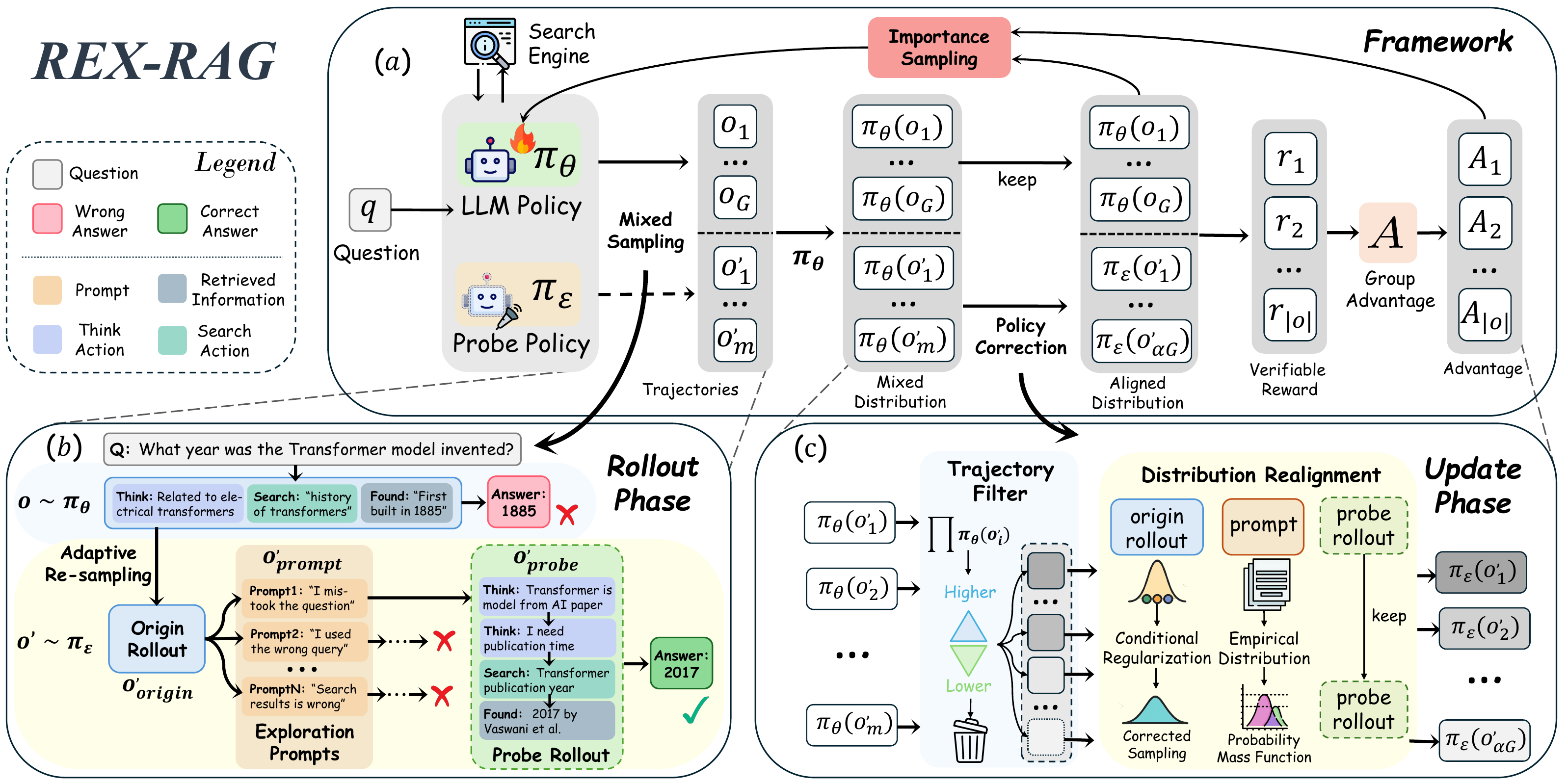}
  \caption{Overview of REX-RAG. (a) Overall framework architecture; (b) Mixed Sampling Strategy in Rollout Phase that combines policy and probe sampling; (c) Policy Correction Mechanism in Update Phase that corrects distribution shift.}
  \label{fig:framework}
\end{figure*}

\section{Method}
\subsection{Preliminary}

\paragraph{RAG Task Formulation} RAG addresses this limitation of LLMs when answering complex questions that require external knowledge beyond their training data. Formally, given a question $q$ and a golden answer $a$ from a dataset $\mathcal{D} = \{(q_i, a_i)\}_{i=1}^{n}$, the LLM alternates between generation and retrieval. At each step, it generates reasoning text or a search query, which is used to retrieve documents $d = \{d_1, d_2, \dots, d_k\}$ from an external knowledge source $\mathcal{R}$ (\textit{e.g.}, a search engine or database), and produces a final answer.

\paragraph{RLVR Enhanced RAG} RLVR extends the RAG framework by integrating retrieval and reasoning into a reinforcement learning loop~\cite{li2025towards}. The learning process is guided by a verifiable reward signal based on an objective correctness criterion, such as exact match.
Formally, for each question-answer pair $(q,y)$, the reward signal $r(q,y)$ provides feedback indicating whether the generated answer satisfies predefined verification criteria. 

\paragraph{GRPO Algorithm}
\label{sec:grpo}
GRPO~\cite{DeepSeekMath} is an emerging RL algorithm for training LLM policies.  Formally, GRPO trains a target policy LLM $\pi_\theta$ using trajectories collected from a previous policy $\pi_{\theta_{\text{old}}}$. The goal is to maximize the expected reward while keeping the learned policy close to a fixed reference policy $\pi_{\mathrm{ref}}$ (e.g., the pre-trained LLM prior to RL fine-tuning), ensuring training stability. For a given query $q$, GRPO generates multiple trajectories through rollouts and computes a normalized reward as the advantage. Moreover, for readability, the descriptions related to GRPO in the main text do not distinguish between $\pi_{\theta_{old}}$ and $\pi_\theta$.

\subsection{REX-RAG Framework}

In this work, we propose REX-RAG, a novel framework that addresses the exploration challenge in RLVR-based RAG through two key innovations. As illustrated in Fig.\ref{fig:framework}, during the Rollout Phase (Fig.~\ref{fig:framework} (b)), a Mixed Sampling Strategy generates diverse trajectories by combining actions from both the target policy $\pi_\theta$ and the probe policy $\pi_\varepsilon$ to escape ``dead ends". In the subsequent Update Phase (Fig.~\ref{fig:framework} (c)), a Policy Correction Mechanism applies importance sampling to correct distribution shifts introduced by mixed sampling,  ensuring stable policy learning while incorporating insights from exploratory rollouts.

\paragraph{RLVR Algorithm} REX-RAG is implemented using GRPO as the underlying reinforcement learning algorithm. As described in Sec. \ref{sec:grpo}, GRPO generates multiple trajectories in Rollout Phase and computes normalized rewards as advantages to update policy parameters in Update Phase.

\paragraph{Structured Interaction Protocol} To facilitate structured interaction between the model and search engine, we adopt the Search-R1 protocol~\cite{Search-r1}, which uses specialized tokens to define different actions during the reasoning process. Specifically, this method use prompt engineering to enables the model to autonomously interact with the search engine through special tokens that trigger different actions. The specific actions are detailed in the Appendix~\ref{app:sec:interaction}.

\paragraph{Reward Function}The reward function is a rule-based reward using exact match. Specifically, the exact match strictly assigns a reward of 1 if the model’s answer exactly matches the golden answer, and 0 otherwise.
\begin{align}
r = \mathrm{EM}(\text{ans}_{\mathrm{pred}}, \text{ans}_{\mathrm{gold}})\text{.}
\end{align}

\subsection{Mixed Sampling Strategy}

The Mixed Sampling Strategy enhances exploration by employing a mixed behavior policy that combines trajectories from both the current policy $\pi_\theta$ and the probe policy $\pi_\varepsilon$, thus, the mixed behavior policy can be formulated as:
\begin{align}
  \mu = \{\pi_\theta, \pi_\varepsilon\}\text{.}
\end{align}

Specifically, the strategy adaptively samples from both policies to maintain exploration diversity. It operates through a two-stages process: first sampling trajectories from the LLM policy, then adaptively performing probe sampling based on the proportion of incorrect paths.

\paragraph{Adaptive Probe Re-sampling} To effectively balance exploration and exploitation, REX-RAG introduces an adaptive probe re-sampling mechanism that dynamically adjusts the degree of exploration based on the observed performance of the current policy. 

The exploration process begins by sampling $n$ trajectories for each question. After collecting the corresponding rewards $\{r_1, r_2, \dots, r_n\}$, where each $r_i \in [0, 1]$, additional exploratory trajectories are sampled in an adaptive manner. Specifically, each trajectory is resampled with probability $p(1 - r_i)$, where $p \in [0, 1]$ is a hyperparameter that controls sampling ratio. This adaptive mechanism encourages more exploration when the policy underperforms and less when it performs well. Consequently, for each question, the expected number of resampled trajectories is given by:
\begin{align}
    m = p \sum_{i=1}^{n} (1 - r_i)\text{.}
\end{align}

\paragraph{Construction of Probe Policy} To enable effective exploration, the probe policy $\pi_\varepsilon$ is constructed using a simple prompt-guided augmentation strategy, which generates exploratory trajectories by injecting exploratory guidance into the original reasoning process.

Each exploratory trajectory $o^{\prime}$ is composed by concatenating three components:
\begin{align}
o^{\prime} = o^{\prime}_{\text{origin}} \oplus o^{\prime}_{\text{prompt}} \oplus o^{\prime}_{\text{probe}}\text{,}
\end{align}
where $\oplus$ denotes sequence concatenation. Specifically:
\begin{itemize}
\item $o^{\prime}_{\text{origin}}$: the original model rollout up to the point where it produces an incorrect or premature answer, preserving the initial reasoning context.
\item $o^{\prime}_{\text{prompt}}$: an exploration prompt sampled from a curated prompt pool $\mathcal{P}$, designed to inject alternative reasoning directions.
\item $o^{\prime}_{\text{probe}}$: a new continuation generated by the target model $\pi_\theta$, conditioned on the modified context.
\end{itemize}

The prompt pool $\mathcal{P}$ is built by rephrasing a comprehensive reflection prompt into $k$ diverse chain-of-thought fragments using GPT-4.5 ~\cite{gpt_4_5}. These fragments represent various reasoning strategies or question reformulations designed to stimulate exploration. The full list of base prompts and their derived fragments are provided in the Appendix~\ref{app:sec:prompt_template}. For more empirical results on different prompts, please refer to Appendix\ref{app:sec:analysis_prompt}.

\subsection{Policy Correction Mechanism}

\paragraph{Distribution Shift Chanllenge} If the mismatch between the behavior policy $\mu=\{\pi_\theta,\pi_\varepsilon\}$ and the target policy $\pi_\theta$ introduced by the mixed sampling strategy is not addressed, model-generated samples are systematically underweighted, whereas tokens from exploration prompts are overweighted. As a result, tokens in inserted spans with negative advantages  may be excessively penalized, potentially falling outside $\pi_\theta$'s support, whereas regions with positive advantages risk entropy collapse due to overly concentrated probabilities. Although GRPO's clipping trick partially addresses these issues, it does not apply during the first update in each training step, leaving the problem unresolved. Fundamentally, using an on-policy estimator in an off-policy setting introduces estimation bias and instability. For detailed mathematical analysis, refer to Appendix~\ref{app:sec:distribution_shift}. To mitigate this, we propose a \emph{Policy Correction Mechanism} (Fig.~\ref{fig:framework} (c)), which reduces distribution shift and gradient bias via two steps: (i) \textit{Trajectory Filtering}, and (ii) \textit{Distribution Realignment}.

\paragraph{Trajectory Filtering} A trajectory filtering mechanism is first introduced to preferentially select rollouts from the probe policy that closely approximate the target policy, thereby mitigating instability and bias. Specifically, trajectories $o'$ are filtered according to their log-likelihood under the current policy $\pi_\theta$, retaining those consistent enough with it. The retention ratio is controlled by a hyperparameter $\alpha$. After filtering, for each question $t$, the retained trajectories are combined with those generated from the target policy:
\begin{align}
  \mathcal{O}_t &= \bigl\{\, o_{i} \mid o_{i} \sim \pi_{\theta} \,\bigr\}_{i=1}^{G} \;\cup\; \bigl\{\, o^{\prime}_{j} \mid o^{\prime}_{j} \sim \pi_{\varepsilon} \,\bigr\}_{j=1}^{\alpha G}\text{.}
\end{align}
\paragraph{Distribution Realignment} Despite the trajectory filtering, a significant distributional mismatch still exists between the mixed behavior policy $\mu$ and the target policy $\pi_\theta$. Specifically, we first define the distribution of the Probe Policy through a principled realignment mechanism. Then, leveraging the theory of multiple importance sampling, we derive a custom GRPO optimization objective to perform parameter updates.

\begin{table*}[htbp]
  \centering
  \caption{Main experimental results on seven QA benchmarks. Best performance is highlighted in \textbf{bold}; the second best is \underline{underlined}. $\heartsuit$ denotes in-domain datasets (trained on), $\diamondsuit$ denotes out-of-domain datasets. All results are exact match accuracy. Additional statistical analysis and significance testing are detailed in the Appendix~\ref{app:sec:analysis_stat}.}
  \label{tab:main_results}
  \begin{tabular}{lcccccccc}
    \toprule
    \multirow{2}{*}{Methods}
      & \multicolumn{3}{c}{General QA}
      & \multicolumn{4}{c}{Multi-Hop QA}
      & \multirow{2}{*}{Avg.} \\
    \cmidrule(lr){2-4} \cmidrule(lr){5-8}
      & NQ$^{\heartsuit}$   & TriviaQA$^{\diamondsuit}$   & PopQA$^{\diamondsuit}$   & HotpotQA$^{\heartsuit}$
      & 2wiki$^{\diamondsuit}$ & Musique$^{\diamondsuit}$    & Bamboogle$^{\diamondsuit}$
      & \\
    \midrule
    \multicolumn{9}{l}{\textbf{Qwen2.5-3B-Base/Instruct}} \\
    \hdashline
    \quad RAG                & 34.8 & 54.4 & 38.7 & 25.5 & 22.6 & 4.7 & 0.8 & 27.0 \\
    \quad IRCoT              & 11.1 & 31.2 & 20.0 & 16.4 & 17.1 & 6.7 & 24.0 & 18.1 \\
    \quad Search-o1          & 23.8 & 47.2 & 26.2 & 22.1 & 21.8 & 5.4 & \textbf{32.0} & 25.5 \\
    \hdashline
    \quad R1-base            & 22.6 & 45.5 & 17.3 & 20.1 & 26.8 & 5.5 & 22.4 & 22.9 \\
    \quad R1-instruct        & 21.0 & 44.9 & 17.1 & 20.8 & 27.5 & 6.0 & 19.2 & 22.4 \\
    \quad Search-R1-base     & \underline{42.1} & \underline{58.3} & \underline{41.3} & 29.7 & 27.4 & 6.6 & 12.8 & 31.2 \\
    \quad Search-R1-instruct & 39.7 & 56.6 & 39.1 & \underline{33.1} & \underline{31.0} & \underline{12.4} & 23.2 & \underline{33.6} \\
    \hdashline
    \quad \textbf{REX-RAG (Ours)}  & \textbf{43.9} & \textbf{60.4} & \textbf{44.2} & \textbf{37.4} & \textbf{39.7} & \textbf{14.5} & \underline{31.2} & \textbf{38.7} \\
    \midrule
    \multicolumn{9}{l}{\textbf{Qwen2.5-7B-Base/Instruct}} \\
    \hdashline
    \quad RAG                & 34.9  & 58.5  & 39.2  & 29.9  & 23.5  & 5.8  & 20.8  & 30.4  \\
    \quad IRCoT              & 22.4  & 47.8  & 30.1  & 13.3  & 14.9  & 7.2  & 22.4  & 23.9  \\
    \quad Search-o1          & 15.1  & 44.3  & 13.1  & 18.7  & 17.6  & 5.8  & 29.6  & 20.6  \\
    \hdashline
    \quad R1-base            & 29.7  & 53.9  & 20.2  & 24.2  & 27.3  & 8.3  & 29.6  & 27.6  \\
    \quad R1-instruct        & 27.0  & 53.7  & 19.9  & 23.7  & 29.2  & 7.2  & 29.3  & 27.1  \\
    \quad Search-R1-base     & 39.5  & 56.0  & 38.8  & 32.6  & 27.0  & 12.5  & 36.0  & 35.0  \\
    \quad Search-R1-instruct & \underline{42.9}  & \underline{62.3}  & \underline{42.7}  & \underline{38.6}  & \underline{34.6}  & \underline{16.2}  & \underline{40.0}  & \underline{39.6}  \\
    \hdashline
    \quad \textbf{REX-RAG (Ours)}  & \textbf{45.5} & \textbf{62.6} & \textbf{44.3} & \textbf{42.2} & \textbf{43.7} & \textbf{19.7} & \textbf{44.8} & \textbf{43.2} \\
    \bottomrule
  \end{tabular}
  \vspace{0.1cm}
\end{table*}

\textbf{Probe Policy Definition} is nontrivial because the probe policy constructs trajectories by augmenting original rollouts with injected prompts and subsequent continuations. To model \(\pi_{\varepsilon}\) accurately, trajectories are decomposed into segments, each modeled individually under $\pi_\varepsilon$ as follows:
\begin{align}
  \pi_{\varepsilon}&(o^{\prime}_{i,t} \mid q_i, o^{\prime}_{i<t}) = \begin{cases}
    \dfrac{\pi_{\theta}(o^{\prime}_{i,t} \mid q_i, o^{\prime}_{i<t})}{z^{1/|o'_{\mathrm{origin}}|}},
    & \text{if } o^{\prime}_{i,t} \in o'_{\mathrm{origin}} \\[1.2em]
    \mathrm{PMF}(o^{\prime}_{i<t}, o^{\prime}_{i,t}),
    & \text{if } o^{\prime}_{i,t} \in o'_{\mathrm{prompt}} \\[1.2em]
    \pi_{\theta}(o^{\prime}_{i,t} \mid q_i, o^{\prime}_{i<t}),
    & \text{if } o^{\prime}_{i,t} \in o'_{\mathrm{probe}}
  \end{cases}\text{.}
\end{align}
\begin{itemize}
\item The prefix segment is treated as sampled from a truncated version of $\pi_\theta$, conditioned on failure, with $z$ representing the empirical failure rate.
\item The prompt segment is deterministically selected, modeled by an empirical probability mass function (PMF) over the prompt pool.
\item The continuation segment is sampled directly from $\pi_\theta$, thus requires no correction.
\end{itemize}

The specific design details and the construction method of the probability mass function based on frequency distribution are provided in the Appendix~\ref{app:sec:ppd}.

\textbf{Multiple Importance Sampling} is then further employed to correct the distributional mismatch between the mixed behavior policy $\mu$, from which data is collected, and the target policy $\pi_{\theta}$, under which the model is optimized.

The importance ratio for action $o_{i,t}$ at time step $t$ within trajectory $i$ is computed according to the balance heuristic ~\cite{BH} as:
\begin{align}
\label{ratio}
\omega_{i,t}
= \frac{(1+\alpha)\,\pi_{\theta}\!\left(o_{i,t} \mid q_i, o_{i,<t}\right)}
{\pi_{\theta}\!\left(o_{i,t} \mid q_i, o_{i,<t}\right) + \alpha\,\pi_{\varepsilon}\!\left(o_{i,t} \mid q_i, o_{i,<t}\right)}\text{.}
\end{align}

The policy is then optimized with the GRPO objective:
\begin{align}
&J_{\text{GRPO}}(\theta)
= \mathbb{E}_{\,q \sim \mathcal{D},\;\{o_i\}_{i=1}^{\mid \mathcal{O}\mid } \sim \mu(\cdot \mid q)} \Bigg[ \frac{1}{\mid \mathcal{O}\mid }\sum_{i=1}^{\mid \mathcal{O}\mid } \frac{1}{|o_i|}\sum_{t=1}^{|o_i|} \notag\\ 
&\quad 
\min\!\Bigg(
\omega_{i,t}\,\hat{A}_{i,t},\; \mathrm{clip}\!\left(
\omega_{i,t},
\,1-\varepsilon,\,1+\varepsilon
\right)\hat{A}_{i,t}
\Bigg)  \notag\\
& \quad - \beta\, D_{\mathrm{KL}}\!\left(\pi_{\theta}\,\|\,\pi_{\mathrm{ref}}\right)
\Bigg]\text{,}
\end{align}
where the behavior policy was updated to a mixture $\mu$, the coefficient multiplying the advantage was updated to the importance ratio in Eq. (\ref{ratio}) rather than simple ratio between the new and old $\pi_\theta$, and the group size was updated to $\mid \mathcal{O} \mid$.

\section{Experiment}

We conduct extensive evaluations of REX-RAG on seven QA benchmarks,  generalizability analysis. Additionals experimental anaylsis on prompts and hyper--parameters are included in the Appendix~\ref{app:sec:analysis}.

\subsection{Experimental Setup}

\begin{table*}[!htbp]
  \centering
  \caption{Ablation study over key components in REX-RAG (Qwen2.5-3B,GRPO).}
  \label{tab:ablation}
  \begin{tabular}{lcccccccc}
    \toprule
    \multirow{2}{*}{Methods}
      & \multicolumn{3}{c}{General QA}
      & \multicolumn{4}{c}{Multi-Hop QA}
      & \multirow{2}{*}{Avg.} \\
    \cmidrule(lr){2-4} \cmidrule(lr){5-8}
      & NQ   & TriviaQA  & PopQA   & HotpotQA
      & 2wiki & Musique    & Bamboogle
      & \\
    \midrule
    \textbf{REX-RAG}  & \textbf{43.9} & \textbf{60.4} & \textbf{44.2} & \textbf{37.4} & \textbf{39.7} & \textbf{14.5} & \textbf{31.2} & \textbf{38.7} \\
    \hdashline
    \qquad   Coarse PPD     & 45.4 & 60.9 & 44.1 & 35.4 & 35.1 & 10.7 & 23.2 & 36.4 \\
    \qquad   w/o IS     & 45.4 & 61.8 & 43.9 & 32.5 & 28.8 & 8.1 & 13.6 & 33.4 \\
    \qquad w/o TF  & 39.7 & 54.2 & 36.6 & 26.0 & 26.4 & 5.5 & 9.6 & 28.2 \\
    \qquad w/o IS\&TF  & 39.5 & 56.1 & 41.5 & 26.6 & 26.0 & 5.3 & 8.8 & 29.1 \\
    \bottomrule
  \end{tabular}
\end{table*}

\paragraph{Datasets} We evaluate REX-RAG on seven QA benchmarks: three general QA datasets NQ~\cite{NQ}, TrivialQA~\cite{TrivialQA}, and PopQA~\cite{PopQA}, together with four Multi-Hop QA datasets HotpotQA~\cite{HotpotQA}, 2WikiMultiHopQA~\cite{2wiki}, Musique~\cite{MuSiQue}, and Bamboogle~\cite{Bamboogle}. In line with earlier studies~\cite{Search-r1,Search-r1-emperical}, we merge the NQ and HotpotQA training sets for REX‑RAG training. The test splits of NQ and HotpotQA are treated as in‑domain evaluations, and the remaining five datasets are used for out‑of‑domain evaluation. For detailed information, please refer to Appendix~\ref{app:sec:details:datasets}.

\paragraph{Baselines}

To evaluate the effectiveness of REX-RAG, we compare it with several baselines, categorized into two groups: (1) non-fine-tuned methods, including Naive RAG~\cite{RAG}, IRCOT~\cite{IRCOT}, and Search-o1~\cite{Search-o1}; and (2) fine-tuned methods, including R1-like~\cite{Deepseek-r1} training using PPO~\cite{PPO} without retrieval and those with retrieval~\cite{Search-r1} using GRPO~\cite{DeepSeekMath}.

\begin{table}[!htbp]
  \centering
  \caption{Algorithm generalizability analysis comparing GRPO and DAPO frameworks on Qwen2.5-3B. Scores represent exact match accuracy (\%) averaged across General QA and Multi-Hop QA categories.}
  \label{tab:algorithm_comparison}
  \begin{tabular}{lccc}
    \toprule
       Methods       &  General QA & Multi-Hop QA & Avg. \\
    \midrule
    \textbf{GRPO} &&\\
    \hdashline
    \quad Search-R1      &     47.2      &    19.1 &  31.2  \\
    \quad REX-RAG      &    49.5        &    30.7  &  38.7 \\
    \midrule
    \textbf{DAPO} &&\\
    \hdashline
    \quad Search-R1      &     50.9      &    22.7  &  34.8 \\
    \quad REX-RAG      &    48.4        &    30.9   &  38.4 \\
    \bottomrule
  \end{tabular}
\end{table}

\paragraph{Implementation Details}

For external search engines, we utilize the December 2018 Wikipedia dump ~\cite{wiki} as our primary data source and employ the E5-base-v2 model ~\cite{E5} as the retriever. During each retrieval step, the top-3 documents returned by the retriever are provided as additional context.

For REX-RAG, we adopt Qwen2.5-3B and Qwen2.5-7B as base models ~\cite{qwen}, using GRPO as the default RL algorithm. The hyperparameters $\alpha$ and $p$ are set to default values of 0.12 and 0.2. For further details on experimental settings, please refer to the Appendix~\ref{app:sec:details}.

For evaluation, we mainly rely on the exact match. Additionally, most of the baseline results in Table \ref{tab:main_results} are taken from Search-R1~\cite{Search-r1, Search-r1-emperical}.

\begin{figure*}[t]
  \centering
  \includegraphics[width=0.96\textwidth]{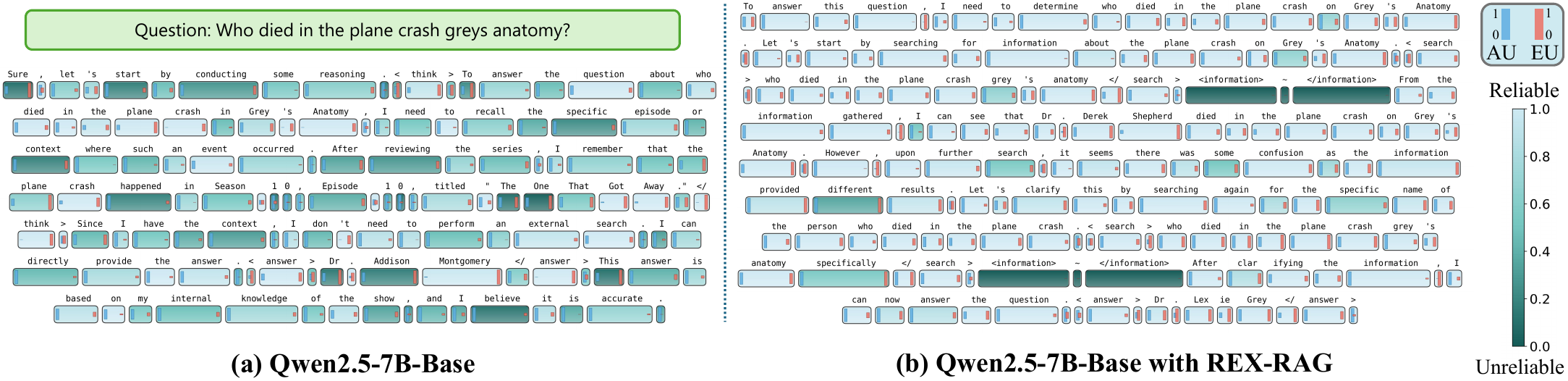}
  \caption{Uncertainty quantification visualization comparing Qwen2.5-7B-Base (left) and REX-RAG (right). Color intensity represents uncertainty levels; Blue bars represent Aleatoric Uncertainty (AU) and orange bars represent Epistemic Uncertainty (EU). REX-RAG demonstrates coherent reasoning with reduced epistemic uncertainty and higher reliability scores.}
  \label{fig:case_study}
\end{figure*}

\subsection{Overall Performance}

Table~\ref{tab:main_results} presents the main experimental results across seven diverse QA benchmarks. REX-RAG demonstrates consistent and substantial improvements over all baseline methods across both model sizes and dataset types.

\paragraph{Performance Gains} REX-RAG achieves significant performance improvements over the strongest baseline (Search-R1-instruct): +5.1\% average improvement on Qwen2.5-3B (38.7\% vs 33.6\%) and +3.6\% on Qwen2.5-7B (43.2\% vs 39.6\%). These gains are particularly pronounced on multi-hop reasoning tasks, where REX-RAG shows +8.7\% improvement on 2Wiki and +4.3\% on HotpotQA for the 3B model, demonstrating the effectiveness of our method.

\paragraph{Out-of-Domain Generalization} REX-RAG also exhibits strong generalization capabilities across out-of-domain datasets. On TriviaQA, PopQA, 2Wiki, MuSiQue, and Bamboogle—none of which were seen during training—REX-RAG consistently outperforms baselines by substantial margins. This suggests that the mixed sampling strategy successfully learns generalizable reasoning patterns rather than overfitting to specific dataset characteristics.

\paragraph{Comparison with Non-Finetuned Methods} REX-RAG significantly outperforms non-finetuned approaches, achieving 13.2\% higher average performance than the best non-finetuned RAG method on 3B models. This demonstrates the value of reinforcement learning for RAG reasoning, while our method further amplifies these benefits.

\subsection{Ablation Studies}

\subsubsection{Ablation on Key Components}

Table~\ref{tab:ablation} presents ablation studies examining the contribution of each component in REX-RAG. We systematically remove or modify key components to understand their individual impact.

\paragraph{Component Analysis} (1) \textbf{Full REX-RAG}: Our complete method achieving 38.7\% average performance. (2) \textbf{Coarse PPD}: Uses a simplified probe policy definition where the first token of inserted prompts receives probability $1/k$, while remaining prompt tokens are assigned probability 1. This coarse approximation leads to a 2.3\% performance drop, demonstrating the importance of accurate probability modeling. (3) \textbf{w/o IS}: Removes importance sampling, treating all trajectories equally during training. This results in a 5.3\% performance degradation. (4) \textbf{w/o TF}: Eliminates trajectory filtering, including all probe-generated trajectories regardless of quality. Performance drops by 10.5\%, showing that quality control is essential for effective exploration. (5) \textbf{w/o IS\&TF}: Removes the entire Policy Correction Mechanism, including IS and TS, essentially reducing to naive trajectory augmentation. This causes a 9.6\% performance drop, confirming that principled distribution correction is crucial for stable learning.

\paragraph{Key Insights} The ablation results reveal several important insights: First, the Policy Correction Mechanism is a critical component, with its removal causing a large performance degradation. Second, trajectory filtering is essential for maintaining training stability—without it, noisy exploratory trajectories significantly harm performance. Third, even coarse probability estimation provides substantial benefits over no correction, though precise modeling yields optimal results. These findings validate the effectiveness of our framework and design choices.

\subsubsection{Algorithm Generalizability}

Table~\ref{tab:algorithm_comparison} demonstrates that REX-RAG's benefits generalize across different reinforcement learning algorithms. When trained with DAPO ~\cite{DAPO} instead of GRPO, REX-RAG maintains substantial improvements over Search-R1 (38.4\% vs 34.8\% average performance), though gains are slightly smaller than with GRPO. This suggests that REX-RAG is algorithm-agnostic and can be integrated with various RL frameworks.
Interestingly, DAPO shows stronger performance on general QA tasks for Search-R1, while GRPO excels on multi-hop reasoning. REX-RAG benefits from both algorithms but shows more consistent improvements with GRPO, likely due to GRPO's group-based advantage estimation being more compatible with our mixed sampling strategy.

\subsection{Case Studies and Visualization}

Fig.~\ref{fig:case_study} presents a comprehensive visualization analysis comparing reasoning trajectories of Qwen2.5-7B-Base and REX-RAG using uncertainty quantification methodology from \textbf{LogTokU~\cite{LogTokU}}. Following the framework, we analyze \textbf{Aleatoric Uncertainty (AU)} representing inherent data randomness and \textbf{Epistemic Uncertainty (EU)} capturing model knowledge gaps through token-level confidence scoring. The visualization demonstrates that REX-RAG achieves universally higher reliability scores for reasoning tokens, with values frequently falling in the 0.6–0.8 range, whereas the baseline exhibits lower reliability (typically in the 0.2–0.4 range). This indicates REX-RAG exhibits superior confidence calibration and more reliable decision-making throughout the reasoning process.

The uncertainty analysis reveals that REX-RAG exhibits high AU combined with low EU, providing evidence that REX-RAG is more exploratory precisely when it possesses relevant knowledge. This behavior demonstrates that REX-RAG's probe policy effectively identifies situations where multiple valid reasoning paths exist (high AU) while maintaining confidence in its knowledge base (low EU), leading to more thorough exploration of the solution space. In contrast, the baseline model shows the opposite pattern with low AU and high EU, indicating overconfidence in limited reasoning paths while lacking awareness of knowledge gaps.

Beyond uncertainty patterns, REX-RAG produces significantly more standardized and coherent output formats compared to the baseline's fragmented and irregular response structures. The visualization clearly shows that REX-RAG maintains logical flow, consistent structure, and systematic reasoning throughout, whereas the base LLM exhibits abrupt transitions, disjointed reasoning, and produces overconfident yet incorrect answer.
This highlights that REX-RAG offers more reliable confidence estimation, coherent reasoning, and overall robustness in RAG reasoning.

\section{Limitation}
We discuss main limitations of our current approach; further details are provided in the Appendix~\ref{app:sec:future_work}.

\paragraph{Limited Exploration Strategy}
Our method relies on fixed-pool prompt insertion, which, though effective, can be improved. Future work could include model-generated prompts, backtracking-based search, or full-path restructuring for more comprehensive exploration.

\paragraph{Computational Overhead}
The mixed sampling strategy introduces additional trajectories due to difficulty assessment followed by resampling. Though more efficient than uniform oversampling, difficulty-predictive sampling could reduce this overhead but remains challenging.

\section{Conclusion}

This work addresses the dead end problem in reinforcement learning-based retrieval-augmented generation, where models become trapped in unproductive reasoning paths during policy optimization. Our REX-RAG framework introduces the Mixed Sampling Strategy and the Policy Correction Mechanism to enable systematic exploration while maintaining training stability. Comprehensive experiments demonstrate consistent improvements over strong baselines, with particularly notable gains on multi-hop reasoning tasks.
Our key contribution lies in providing a principled approach to exploration in LLM reasoning systems through importance sampling-based distributional correction. 
This insight may offers a practical solution for improving retrieval-augmented generation systems and provides a new exploration perspective for LLM reinforcement learning.

\bibliography{aaai2026}

\appendix

\clearpage
\setcounter{secnumdepth}{2}
\section*{Contents}
\begin{description}
    \item [\textbf{A}] \textbf{Additional Experimental Analysis} ...........................  \pageref{app:sec:analysis}
        \begin{description}
            \item [A.1] Analysis of Hyper-parameters ............................. \pageref{app:sec:analysis_hyper}
            \item [A.2] Analysis of Exploration Prompt .......................... \pageref{app:sec:analysis_prompt}
            \item [A.3] Statistical Analysis and  Significance Test ........... \pageref{app:sec:analysis_stat}
        \end{description}
    \item [\textbf{B}] \textbf{Mathematical Formulations and Derivations} ...........\pageref{app:sec:formulas}
        \begin{description}
            \item[B.1] GRPO Algorithm .................................................  \pageref{app:sec:grpo}
            \item[B.2] Distribution Shift .................................................  \pageref{app:sec:distribution_shift}
            \item[B.3] Probe Policy Definition .......................................  \pageref{app:sec:ppd}
            \item[B.4] Coefficient for Importance Sampling  ..................  \pageref{app:sec:coe}
        \end{description}
    \item [\textbf{C}] \textbf{Experimental Implementation Details} ..................... \pageref{app:sec:details}
        \begin{description}
            \item [C.1] Baseline Methods ................................................ \pageref{app:sec:details:comparison_methods}
            \item [C.2] Dataset Descriptions ............................................ \pageref{app:sec:details:datasets}
            \item [C.3] Computational Environment and Infrastructure .. \pageref{app:sec:details:env}
            \item [C.4] Hyper-parameter Configuration and Tuning ....... \pageref{app:sec:details:params}
        \end{description}
    \item [\textbf{D}] \textbf{Limitations, Discussion, and Future Work}............... \pageref{app:sec:future_work}
    \item [\textbf{E}] \textbf{Structured Search Interaction Protocol}.....................\pageref{app:sec:interaction}
        \begin{description}
            \item [E.1] Special Tokens .....................................................
            \pageref{app:sec:special_tokens}
            \item [E.2] System Prompt .................................................... \pageref{app:sec:system_prompt}
        \end{description}
    \item [\textbf{F}] \textbf{Prompt Templates and Examples}...............................\pageref{app:sec:prompt_template}
\end{description}

\newpage

\section{Additional Experimental Analysis}
\label{app:sec:analysis}

\subsection{Analysis of Hyper-parameters}
\label{app:sec:analysis_hyper}

REX-RAG introduces a hyperparameter $p$ that controls the number of additionally sampled trajectories. As shown in the Table~\ref{tab:sampling_comparison}, sampling only an extra 12\% of trajectories yields a substantial performance improvement over Search-R1. By contrast, Search-R1 attains only a negligible gain even when using 20\% more trajectories, highlighting the superior sample efficiency of REX-RAG. Moreover, we observe a positive correlation between model performance and the resampling parameter $p$; with 20\% additional sampling, the improvement becomes even more pronounced. This property allows practitioners to flexibly trade off performance gains against computational cost according to their specific needs and resource constraints.

\begin{table}[ht]
  \centering

  \begin{tabular}{lccc}
    \toprule
       Sampling Strategy      &  General & Multi-Hop & Avg. \\
    \midrule
    \multicolumn{3}{l}{\textbf{Search-R1}} \\
    \hdashline
    5 rollouts (+0\%)      &     47.2      &    19.1   & 31.2  \\
    6 rollouts (+20\%)     &    47.6        &    19.1  & 31.3   \\
    \midrule
    \multicolumn{3}{l}{\textbf{REX-RAG}} \\
    \hdashline
    5.6 (+12\% $\leftarrow$ 12\%)     &    48.7        &    23.4    & 34.2 \\
    5.6 (+12\% $\leftarrow$ 20\%)      &     49.5      &    30.7  & 38.7   \\
    \bottomrule
  \end{tabular}
  \caption{Impact of trajectory sampling strategies on performance. Expected rollout counts shown for REX-RAG under maximum resampling scenarios (all initial outputs incorrect).}
  \label{tab:sampling_comparison}
\end{table}

\subsection{Analysis of Exploration Prompt}
\label{app:sec:analysis_prompt}

As shown in the Table~\ref{tab:prompt_comparision}, we examined how varying exploration prompts affects model performance. With five prompts, we observe modest improvements on General QA and Multi-Hop QA. However, when expanding from five to thirty prompts, REX-RAG achieves a substantial performance gain relative to Serach-R1. These results indicate that the REX-RAG framework exhibits strong scalability, rather than merely benefiting from a small set of specially selected prompts.

\begin{table}[ht]
  \centering

  \begin{tabular}{lccc}
    \toprule
       Sampling Strategy      &  General& Multi-Hop & Avg. \\
    \midrule
    Search-R1     &      47.2       &     19.1      &    31.2    \\
    REX-RAG(5 Prompts)      &     48.3      &    20.0   & 32.1  \\
    REX-RAG(30 Prompts)      &     49.5      &    30.7   & 38.7  \\
    \bottomrule
  \end{tabular}
  \caption{Impact of Number of Exploration Prompt}
  \label{tab:prompt_comparision}
\end{table}

\subsection{Statistical Analysis and  Significance Test}
\label{app:sec:analysis_stat}

Given that Exact Match is a binary evaluation metric, we adopt the McNemar test to determine whether the performance differences observed in the ablation study constitute statistically significant improvements or degradations. As shown in Table~\ref{tab:ablation}, we evaluate a total of five models. In this subsection, we first rank the models by their Average scores in descending order and then perform pairwise comparisons between successive models.

Each numerical value in the Table~\ref{tab:stat} represents the p-value corresponding to the statistical test of the alternative hypothesis, evaluating the difference between the two models across various benchmark.

As shown in Table~\ref{tab:stat}, the majority of the test results are significant (p-value $<$ 0.05). While the results on a few individual benchmarks are not statistically significant, this does not affect the overall conclusions presented in the main text.

\begin{table*}[bhtp]
  \centering
  \caption{Significance Test over key components in REX-RAG (Qwen2.5-3B,GRPO). Overall represents the results of the tests conducted on seven benchmarks. The rest are the test results obtained on each benchmark.}
  \label{tab:stat}
  \begin{tabular}{lcccccccc}
    \toprule
    \multirow{2}{*}{Alternative Hypothesis}
      & \multicolumn{3}{c}{General QA}
      & \multicolumn{4}{c}{Multi-Hop QA}
      & \multirow{2}{*}{Overall} \\
    \cmidrule(lr){2-4} \cmidrule(lr){5-8}
      & NQ   & TriviaQA  & PopQA   & HotpotQA
      & 2wiki & Musique    & Bamboogle
      & \\
    \midrule
    REX-RAG $\neq$ Coarse PPD & $1e{-9}$ & $5e{-15}$ & $4e{-9}$ & $1e{-50}$ & $1e{-74}$ & $1e{-10}$ & $2e{-2}$ & $5e{-144}$ \\
    Coarse PPD  $\neq$ w/o IS   & $9e{-1}$ & $1e{-3}$ & $4e{-1}$ & $2e{-10}$ & $8e{-55}$ & $1e{-5}$ & $2e{-2}$ & $3e{-33}$ \\
     w/o IS  $\neq$w/o IS\&IF & $4e{-20}$ & $4e{-103}$ & $6e{-125}$ & $3e{-52}$ & $2e{-9}$ & $1e{-8}$ & $2e{-1}$ & $7e{-250}$ \\
   w/o IS\&IF $\neq$ w/o TF  & $7e{-1}$  & $3e{-7}$  & $1e{-53}$  & $1e{-1}$  & $2e{-1}$  & $7e{-1}$  & 1 & $1e{-24}$  \\
    \bottomrule
  \end{tabular}
\end{table*}

\section{Mathematical Formulations and Derivations}
\label{app:sec:formulas}

\subsection{GRPO Algorithm}
\label{app:sec:grpo}

GRPO~\cite{DeepSeekMath} is a reinforcement-learning algorithm for aligning large language models that removes the value/critic network by computing \emph{group-relative} advantages across multiple sampled outputs for the same prompt. The baseline is the group’s average reward, and policy updates are additionally regularized by a KL term to a frozen reference model.

For each prompt $q$, sample a group of $G$ outputs $\{o_i\}_{i=1}^{G}$ from the old policy $\pi_{\theta_{\mathrm{old}}}$.
Define the likelihood ratio
$\rho_{i,t} = \frac{\pi_{\theta}(o_{i,t}\mid q, o_{i,<t})}{\pi_{\theta_{\mathrm{old}}}(o_{i,t}\mid q, o_{i,<t})}$.
GRPO maximizes:
\begin{align}
J_{\text{GRPO}}(\theta) &= \notag \\[-0.4em]
\mathbb{E}_{q\{o_i\}\sim \pi_{\theta_{\mathrm{old}}}}
&\left[
\frac{1}{G}\sum_{i=1}^{G}\frac{1}{|o_i|}
\sum_{t=1}^{|o_i|}
\big(
\min\!\big(\rho_{i,t}\,\widehat{A}_{i,t},\,
\operatorname{clip}(\rho_{i,t},\varepsilon)\,\widehat{A}_{i,t}\big)
\big)
\right] \notag \\
&- \beta\,\widehat{D}_{\mathrm{KL}}\!\left(\pi_{\theta}\,\Vert\,\pi_{\mathrm{ref}}\right),
\end{align}
where $\varepsilon$ is the PPO clipping parameter and $\beta$ controls KL regularization to the reference policy $\pi_{\mathrm{ref}}$.

\paragraph{Unbiased per-token KL estimator}
GRPO uses a per-token estimator of the forward KL:
\begin{align}
\widehat{D}_{\mathrm{KL}}\!\left(\pi_{\theta}\Vert\pi_{\mathrm{ref}}\right)
=& \mathbb{E}_{(q,o_{i,<t})\sim \pi_{\theta}}
\left[
\frac{\pi_{\mathrm{ref}}(o_{i,t}\mid q,o_{i,<t})}{\pi_{\theta}(o_{i,t}\mid q,o_{i,<t})}
- \right. \notag \\
&\left.
\log\frac{\pi_{\mathrm{ref}}(o_{i,t}\mid q,o_{i,<t})}{\pi_{\theta}(o_{i,t}\mid q,o_{i,<t})}
-1
\right].
\end{align}

\paragraph{Outcome supervision} Let $r_{\phi}$ denote a reward scoring each output. For a fixed $q$, obtain rewards $r=\{r_i\}_{i=1}^{G}$, one per output $o_i$.
Compute the group mean and standard deviation
\[
\mu_r=\frac{1}{G}\sum_{i=1}^{G}r_i,
\qquad
\sigma_r=\operatorname{std}(r_1,\dots,r_G).
\]
Normalize each reward
$\tilde r_i=\frac{r_i-\mu_r}{\sigma_r}$,
and assign a constant advantage to all tokens in $o_i$:
\begin{align}
\widehat{A}_{i,t}=\tilde r_i,\qquad \forall\,t\in\{1,\dots,|o_i|\}.
\end{align}

\subsection{Distribution Shift}
\label{app:sec:distribution_shift}

For the sake of analytical simplicity, we disregard the clipping technique and the KL-divergence regularization term in GRPO. If we intend to employ data drawn from the mixture policy $\mu$ to optimize the target policy $\theta$, the unbiased gradient is given by:

\begin{align}
\nabla_\theta J(\theta)
= \mathbb{E}_{q,\{o_i\}\sim\mu}\!\!
\left[
\frac{1}{G}
\sum_{i=1}^{G}
\frac{1}{\lvert o_i\rvert}
\sum_{t=1}^{\lvert o_i\rvert}
\hat A_{i,t}\,
\nabla_\theta \frac{\pi_{\theta}(o_{i,t}\mid q,o_{i,<t})}{\mu(o_{i,t}\mid q,o_{i,<t})}
\right].
\end{align}

If, instead, we apply no corrective procedure and directly use the data collected under the mixture policy $\mu$ to optimize $\theta$, the gradient we actually compute becomes:

\begin{align}
\tilde{g}(\theta)
= \mathbb{E}_{q,\{o_i\}\sim\mu}\!\!
\left[
\frac{1}{G}
\sum_{i=1}^{G}
\frac{1}{\lvert o_i\rvert}
\sum_{t=1}^{\lvert o_i\rvert}
\hat A_{i,t}\,
\nabla_\theta \frac{\pi_{\theta}(o_{i,t}\mid q,o_{i,<t})}{\pi_{\theta_{\mathrm{old}}}(o_{i,t}\mid q,o_{i,<t})}
\right].
\end{align}

Subtracting the two importance ratios yields the bias:

\begin{align}
    \Delta_{i,t} &= \tilde{\rho}_{i,t} - \rho_{i,t} \notag \\
                 &= \frac{\pi_{\theta, i,t}}{\pi_{\theta_{\mathrm{old}}, i,t}} - \frac{\pi_{\theta, i,t}}{\mu_{i,t}} \notag \\ 
                 &= \frac{\pi_{\theta, i,t}}{\mu_{i,t}} \cdot \left( \frac{\mu_{i,t} - \pi_{\theta_{\mathrm{old}}, i,t}}{\mu_{i,t}} \right) \notag \\ 
                 &= \tilde{\rho}_{i,t} \left( 1 - \frac{\pi_{\theta_{\mathrm{old}}, i,t}}{\mu_{i,t}} \right).
\end{align}

In this expression, the first factor, $\tilde{\rho}_{i,t}$, is strictly positive and can therefore be ignored. Focusing on the sign of the second factor, we observe that for tokens generated freely by the model, $\mu$ comprises both $\pi_{\theta}$ and $\pi_{\epsilon}$, where $\pi_{\epsilon}$ is defined only along erroneous trajectories. Consequently, $\mu$ is smaller than $\pi_{\theta}$, rendering the second factor negative. Thus, for tokens sampled freely by the model, the importance ratio is biased downward, leading to systematic underestimation.

Conversely, for the segments inserted by the probe policy, the second factor is positive, conferring a systematic up-weighting. This persistent high weighting can drive the probabilities of tokens with negative advantages to decline rapidly, potentially pushing them outside the support of the policy model. Tokens with positive advantages, on the other hand, may experience rapid probability increases, thereby squeezing the probabilities of alternative tokens and inducing severe entropy collapse.

\subsection{Probe Policy Definition}
\label{app:sec:ppd}

For the Probe Policy, we partition the procedure into three components according to their ordering relative to the inserted prompt: (1) the segment of the model rollout up to the point of failure; (2) the inserted prompt; and (3) the subsequent trajectory obtained by conditioning on the erroneous reasoning path and the prompt as context.

\begin{align}
  \pi_{\varepsilon}&(o^{\prime}_{i,t} \mid q_i, o^{\prime}_{i<t}) = \begin{cases}
    \dfrac{\pi_{\theta}(o^{\prime}_{i,t} \mid q_i, o^{\prime}_{i<t})}{z^{1/|o'_{\mathrm{origin}}|}},
    & \text{if } o^{\prime}_{i,t} \in o'_{\mathrm{origin}} \\[1.2em]
    \mathrm{PMF}(o^{\prime}_{i<t}, o^{\prime}_{i,t}),
    & \text{if } o^{\prime}_{i,t} \in o'_{\mathrm{prompt}} \\[1.2em]
    \pi_{\theta}(o^{\prime}_{i,t} \mid q_i, o^{\prime}_{i<t}),
    & \text{if } o^{\prime}_{i,t} \in o'_{\mathrm{probe}}
  \end{cases}\text{.}
\end{align}

First, for the segment of the model rollout up to the point where an error occurs, our aim is to model the region of the original policy distribution that gives rise to failures. Within the set of all trajectories that can be sampled from the original distribution, we approximate this subset using $z$, defined as the fraction of erroneous trajectories among those sampled at the current step. This yields a distribution that is truncated relative to the original policy. To make this subset of trajectories a valid probability distribution—that is, to let “the probability mass of these trajectories fill the entire space”—we renormalize it. Accordingly, we divide the probability of each token by $z^{1/\lvert o'_{\mathrm{origin}}\rvert}$ as a simple sequence-level normalization.

For the inserted prompt part, we define it based on the frequency distribution. The method induces a discrete vocabulary via a tokenizer and builds a nonparametric next-token model by aggregating, for each observed prefix $p$, the multiset of successor tokens from the corpus. Each prefix is mapped to a count vector over the vocabulary; the probability mass function is the normalized frequency. Conceptually, this is an unsmoothed, memory-based (variable-length $n$-gram) estimator that returns the empirical conditional distribution of the next token given $p$, assigning zero mass to unseen events. Specifically, the construction algorithm is as shown in Algorithm~\ref{alg:build_pmf_model}.

For the last part, since we do not impose any restrictions on the sampling of these parts, we directly use the probability of the original policy model as the probability of the probe policy.

\begin{algorithm}[ht]
  \caption{PMF Construction via Frequency Distribution}
  \label{alg:build_pmf_model}
  \textbf{Input}: tokenizer $\mathcal{T}$; prompt set $\mathcal{P}=\{s_1,\dots,s_m\}$\\
  \textbf{Output}: function $\mathrm{PMF}(p,x)$
  \begin{algorithmic}[1]
    \STATE $K \gets \bigl\{\mathcal{T}(s)\;|\;s\in\mathcal{P}\bigr\}$ \COMMENT{tokenise every prompt}
    \STATE $V \gets$ unique tokens in $K$  \COMMENT{initialize vocabulary}
    \FORALL{$k \in K$}  \FORALL{$i < |k|-1$}
    \STATE $p \gets k_{0:i}$
    \STATE $C[p] \gets \mathbf{0}_{|V|}$ \COMMENT{initialize frequency distribution}
    \ENDFOR\ENDFOR
    \FORALL{$k \in K$}  \FORALL{$i < |k|-1$}
    \STATE $p \gets k_{0:i}$ ;\; $x \gets k_{i+1}$
    \STATE $C[p][V.\text{index(x)}] \mathrel{+}= 1$
    \ENDFOR\ENDFOR
    \STATE
    \STATE \textbf{define} $\mathbf{function}\;\mathrm{PMF}(p,x)$
    \STATE \hspace{1em}$\text{counts} \gets C[p]$
    \STATE \hspace{1em}\textbf{return} $\dfrac{\text{counts}[V.\text{index(x)}]}{\sum \text{counts}}$
    \STATE
    \STATE \textbf{return} $\mathrm{PMF}$ \COMMENT{exposes the query function to the caller}
  \end{algorithmic}
\end{algorithm}

\subsection{Coefficient for Importance Sampling}
\label{app:sec:coe}

Let the goal be to estimate the policy gradient using a mixed policy $\mu = \{\pi_{\theta},\pi_{\epsilon}\}$. During sampling, a fraction of $\frac{1}{1+\alpha}$ of the trajectories come from $\pi_{\theta}$, while a fraction of $\frac{\alpha}{1+\alpha}$ of the trajectories come from $\pi_{\epsilon}$:
\begin{align}
& c_{\theta}=\frac{1}{1+\alpha},\qquad c_{\epsilon}=\frac{\alpha}{1+\alpha}.
\end{align}
Under the \emph{balance heuristic}~\cite{BH}, the weight is
\begin{align}
\hat{\omega}_i(x)=\frac{c_i\,p_i(x)}{\sum_{j}c_jp_j(x)}.
\end{align}
Substitute the variables into it respectively, and we can obtain the Importance ratio for estimating the policy gradient of Multiple Importance Sampling:
\begin{align}
    \omega
= \frac{(1+\alpha)\,\pi_{\theta}}
{\pi_{\theta} + \alpha\,\pi_{\varepsilon}}\text{.}
\end{align}

\section{Experimental Implementation Details}
\label{app:sec:details}

\subsection{Baseline Methods}
\label{app:sec:details:comparison_methods}
We evaluate REX-RAG against two categories of baselines: (1) non-fine-tuned methods, including Naive RAG~\cite{RAG}, IRCOT~\cite{IRCOT}, and Search-o1~\cite{Search-o1}; and (2) fine-tuned methods, including R1-like~\cite{Deepseek-r1} trained with PPO~\cite{Search-r1} (with and without retrieval) using GRPO~\cite{DeepSeekMath}.

\textbf{Naive RAG}~\cite{RAG} is the standard retrieval-augmented generation approach that retrieves documents using dense passage retrieval and generates answers conditioned on both the query and the retrieved context. It employs a bi-encoder architecture and marginalizes over retrieved documents during generation, enabling dynamic access to external knowledge and reducing hallucination in knowledge-intensive tasks.

\textbf{IRCOT}~\cite{IRCOT} interleaves reasoning and retrieval steps, alternating between generating intermediate reasoning steps and retrieving new information. This few-shot prompting approach enables step-wise information gathering and supports multi-hop reasoning by refining retrieval based on the evolving reasoning chain.

\textbf{Search-o1}~\cite{Search-o1} enhances LLM reasoning by integrating web search. It uses multi-step reasoning to analyze queries, formulate searches, and synthesize results. Iterative search-query reformulation and result ranking improve retrieval quality. The approach relies on chain-of-thought reasoning to generate comprehensive answers using diverse sources.

\textbf{R1-like Training}~\cite{Deepseek-r1} employs RLHF via PPO to fine-tune LLMs for reasoning tasks without retrieval. Following DeepSeek-R1, it includes supervised reasoning trace training, reward modeling, and PPO optimization. This pipeline enhances reasoning quality using curated datasets and human feedback, serving as a strong non-retrieval baseline.

\textbf{Search-R1}~\cite{Search-r1} extends R1-style training by integrating retrieval actions into the policy optimization process using GRPO. It jointly optimizes reasoning and retrieval quality, with rewards based on final answer accuracy and coherence. Retrieval is treated as part of the trajectory, allowing the model to learn effective information-seeking strategies. This serves as a strong prior baseline for evaluating the improvements brought by our proposed policy realignment mechanisms.

\subsection{Dataset Descriptions}
\label{app:sec:details:datasets}

We evaluate REX-RAG on seven QA benchmarks: three general QA datasets NQ~\cite{NQ}, TrivialQA~\cite{TrivialQA}, and PopQA~\cite{PopQA}, together with four Multi-Hop QA datasets HotpotQA~\cite{HotpotQA}, 2WikiMultiHopQA~\cite{2wiki}, Musique~\cite{MuSiQue}, and Bamboogle~\cite{Bamboogle}. In line with earlier studies~\cite{Search-r1,Search-r1-emperical}, we merge the NQ and HotpotQA training sets for REX‑RAG training. The test splits of NQ and HotpotQA are treated as in‑domain evaluations, and the remaining five datasets are used for out‑of‑domain evaluation.

\textbf{Natural Questions (NQ)}~\cite{NQ} is a large-scale dataset featuring real Google Search queries paired with Wikipedia passages containing the answers. It includes over 300K naturally occurring questions, each annotated with both a long answer (usually a paragraph) and a short answer (typically a phrase). NQ reflects realistic information-seeking behavior across diverse topics such as history, science, and current events, with varying complexity. We use it as an in-domain benchmark, as it contributes to REX-RAG’s training.

\textbf{TriviaQA}~\cite{TrivialQA} is a reading comprehension dataset containing over 95K question-answer pairs sourced from trivia websites and paired with evidence documents from Wikipedia and the web. Not all documents are guaranteed to contain the answer, requiring models to perform effective retrieval. The questions emphasize factual knowledge, making the dataset ideal for evaluating retrieval-augmented systems.

\textbf{PopQA}~\cite{PopQA} targets popular factual questions about widely known topics such as celebrities, movies, and sports events. It evaluates models’ ability to answer questions about current and trending topics that may not appear in training corpora, highlighting the importance of real-time retrieval for up-to-date knowledge.

\textbf{HotpotQA}~\cite{HotpotQA} is a multi-hop QA dataset with over 113K Wikipedia-based examples, where each question requires reasoning across at least two paragraphs. It includes bridge and comparison questions and provides supporting facts. As an in-domain benchmark, it plays a key role in evaluating REX-RAG’s multi-hop reasoning performance.

\textbf{2WikiMultiHopQA}~\cite{2wiki} extends multi-hop QA by requiring reasoning over two Wikipedia articles using varied operations like numerical, logical, and compositional reasoning. Each question involves exactly two hops and is annotated with reasoning paths and supporting evidence, facilitating fine-grained evaluation of multi-step reasoning.

\textbf{MuSiQue}~\cite{MuSiQue} focuses on compositional multi-hop reasoning across multiple documents. Questions often involve temporal or relational reasoning and require synthesizing scattered information. It includes both answerable and unanswerable questions, testing models’ ability to detect insufficient context.

\textbf{Bamboogle}~\cite{Bamboogle} is a challenging multi-hop QA benchmark designed to stress-test reasoning capabilities. Questions involve complex inference steps, including temporal and causal reasoning, often under ambiguous or incomplete information. It highlights the limitations of current QA systems and the need for more advanced reasoning strategies.

\subsection{Computational Environment and Infrastructure}
\label{app:sec:details:env}

All experiments in this study were conducted on a cluster of 8 NVIDIA A800 80GB GPUs, providing the computational resources necessary for large-scale reinforcement learning training and evaluation of retrieval-augmented generation systems.

\textbf{Reinforcement Learning Framework.} We implemented our REX-RAG training pipeline using VERL~\cite{verl2024}, an open-source distributed reinforcement learning framework developed by ByteDance for efficient large language model training. VERL is specifically designed to handle the computational challenges of RLHF at scale, providing optimized implementations of policy optimization algorithms such as PPO and GRPO. 

\textbf{Retrieval Infrastructure.} Our retrieval system is built upon FAISS (Facebook AI Similarity Search)~\cite{faiss} for efficient similarity search and indexing. We employ the E5 embedding model~\cite{E5} to encode both queries and documents into dense vector representations, enabling semantic similarity matching for retrieval operations. The knowledge base consists of Wikipedia passages from the DPR corpus~\cite{wiki}, specifically the Wiki-18 dataset. The entire retrieval system is deployed using FastAPI.

\textbf{Data Processing and Evaluation Pipeline.} For data preprocessing, evaluation metrics computation, and baseline comparisons, we adopted the experimental framework from Search-R1~\cite{Search-r1}. This includes standardized data loading procedures, question-answer pair formatting, retrieval corpus preparation, and evaluation protocols that ensure fair comparison across different methods. The Search-R1 framework provides implementations for computing exact match accuracy for multi-hop reasoning evaluation.

\textbf{Prompt Generation and Template Management.} We utilized GPT-4.5 for generating high-quality prompts and reasoning templates used throughout our experiments. This mainly includes the generation of exploration prompts for policy training, as shown in Appendix~\ref{app:sec:prompt_template}. 

\subsection{Hyper-parameter Configuration and Tuning}
\label{app:sec:details:params}

Our hyperparameter configuration strategy primarily focused on tuning algorithm-agnostic parameters that optimize GPU computational performance, particularly those related to macro batch size and GPU utilization settings. This approach ensures efficient resource utilization while maintaining training stability. For all other hyperparameters not directly related to computational performance, we maintained consistency with the Search R1 baseline configuration to ensure fair comparison and reproducibility.
Table~\ref{tab:hyperparams} presents the key hyperparameters used in our REX-RAG implementation. 

\begin{table}[ht]
  \caption{Primary hyperparameters used by REX-RAG. Performance-related parameters were tuned for optimal GPU utilization, while other parameters follow Search R1 baseline configuration.}
  \label{tab:hyperparams}
  \centering
  \begin{tabular}{lcc}
    \toprule
    \textbf{Category} & \textbf{Hyperparameter} & \textbf{Value} \\
    \midrule
    \multirow{6}{*}{\textbf{Performance}} & Training Batch Size & 512 \\
    & Mini Batch Size & 256 \\
    & Max Token Length & 24,000 \\
    & GPU Memory Utilization & 0.8 \\
    & Max Batched Tokens & 8,192 \\
    & Max Sequences per Batch & 1,024 \\
    \midrule
    \multirow{5}{*}{\textbf{Training}} & Actor Learning Rate & $1 \times 10^{-6}$ \\
    & Warmup Steps Ratio & 0.285 \\
    & Weight Decay & 0.01 \\
    & PPO Epochs & 1 \\
    \midrule
    \multirow{4}{*}{\textbf{Policy}} & Clip Ratio & 0.2 \\
    & KL Coefficient & 0.001 \\
    & Use Dynamic Batch Size & True \\
    \midrule
    \multirow{4}{*}{\textbf{Generation}} & Max Search Turns & 5 \\
    & Response Length & 500 \\
    & Temperature & 1.0 \\
    & Top-p Value & 1.0 \\
    \bottomrule
  \end{tabular}
\end{table}

\begin{figure*}[ht]
\caption{Complete prompt template and example interaction for the structured search protocol}
\label{app:fig:prompt_template}
\centering
\begin{tcolorbox}[
    enhanced,
    colback=blue!5!white,
    colframe=blue!75!black,
    title={\textbf{Structured Search Interaction Protocol}},
    fonttitle=\bfseries,
    width=\textwidth,
    sidebyside,
    sidebyside align=top,
    lefthand width=0.48\textwidth,
    righthand width=0.48\textwidth,
    bicolor,
    colbacklower=green!5!white,
    lower separated=false
]
\textbf{Prompt Template:}
\vspace{0.3em}

\small
\textit{Answer the given question. You must conduct reasoning inside \texttt{<think>} and \texttt{</think>} first every time you get new information. After reasoning, if you find you lack some knowledge, you can call a search engine by \texttt{<search>} query \texttt{</search>}, and it will return the top searched results between \texttt{<information>} and \texttt{</information>}. You can search as many times as you want. If you find no further external knowledge needed, you can directly provide the answer inside \texttt{<answer>} and \texttt{</answer>} without detailed illustrations. For example, \texttt{<answer>} xxx \texttt{</answer>}. Question: [question]}

\tcblower

\textbf{Example Interaction:}
\vspace{0.3em}

\footnotesize
\textbf{Question:} What is the capital of France and when was it founded?

\texttt{<think>}
I need to find information about Paris, the capital of France, and its founding date.
\texttt{</think>}

\texttt{<search>}
Paris France capital founding date history
\texttt{</search>}

\texttt{<information>}
Paris is the capital of France. The city was founded in the 3rd century BC by the Parisii tribe...
\texttt{</information>}

\texttt{<think>}
Based on the retrieved information, I now have the answer to both parts of the question.
\texttt{</think>}

\texttt{<answer>}
The capital of France is Paris, which was founded in the 3rd century BC.
\texttt{</answer>}
\end{tcolorbox}
\end{figure*}

The performance-oriented hyperparameters in the first category were specifically tuned to optimize computational efficiency on our hardware configuration. The training batch size of 512 and PPO mini batch size of 256 were selected to maximize throughput while maintaining gradient stability. The GPU memory utilization of 0.8 ensures efficient memory usage without risking out-of-memory errors during training. The token length and sequence limits were configured to balance between accommodating longer reasoning chains and maintaining computational feasibility.

To support deeper reasoning and allow recovery from failed attempts, we increased the Max Search Turns from 2 (as used in Search-R1) to 5. This extension enables the model to conduct further retrieval after initial errors, which is essential for our exploration-driven method. Notably, we kept the overall maximum token budget unchanged, ensuring that this change does not introduce significant additional computational overhead.

All remaining hyperparameters, including learning rates, regularization coefficients, and generation parameters, were kept consistent with the Search-R1 baseline to ensure that performance improvements can be attributed to our proposed REX-RAG methodology rather than hyperparameter optimization advantages.

\begin{table*}[ht]
\centering
\caption{Complete collection of revision prompts used in REX-RAG for triggering self-reflection during reasoning}
\label{app:tab:revision_prompts}
\footnotesize
\begin{tabular}{@{}p{0.6cm}p{7.2cm}p{0.6cm}p{7.2cm}@{}}
\toprule
\textbf{ID} & \textbf{Revision Prompt Text} & \textbf{ID} & \textbf{Revision Prompt Text} \\
\midrule
0 & \texttt{<think>} Perhaps I've overlooked critical points or slipped up in my logic. & 15 & \texttt{<think>} Concerned I might have overlooked key aspects or made subtle errors. \\
\addlinespace[2pt]
1 & \texttt{<think>} I wonder if vital information escaped my notice or if I made an error. & 16 & \texttt{<think>} I might have unintentionally ignored essential details or misunderstood something. \\
\addlinespace[2pt]
2 & \texttt{<think>} There might be key gaps in my understanding or errors in reasoning. & 17 & \texttt{<think>} Revisiting carefully, perhaps errors or oversights went unnoticed earlier. \\
\addlinespace[2pt]
3 & \texttt{<think>} It's possible I've missed something important or misunderstood crucial details. & 18 & \texttt{<think>} Maybe important points slipped my attention, or I made a miscalculation. \\
\addlinespace[2pt]
4 & \texttt{<think>} I suspect errors crept in, or essential points went unnoticed. & 19 & \texttt{<think>} It's likely I've overlooked something crucial or stumbled in logic. \\
\addlinespace[2pt]
5 & \texttt{<think>} Maybe I've misjudged something important or neglected key facts. & 20 & \texttt{<think>} Reflecting, I could've missed critical clues or made errors in judgment. \\
\addlinespace[2pt]
6 & \texttt{<think>} Reflecting now, I might have overlooked critical data or erred somewhere. & 21 & \texttt{<think>} Possibly, I misunderstood something fundamental or missed key evidence. \\
\addlinespace[2pt]
7 & \texttt{<think>} Possibly, I've missed significant insights or made a mistake. & 22 & \texttt{<think>} Concerned about potential unnoticed mistakes or overlooked essential details. \\
\addlinespace[2pt]
8 & \texttt{<think>} I'm sensing a gap or error might be present in my recent reasoning. & 23 & \texttt{<think>} Perhaps my earlier step wasn't entirely accurate or lacked vital points. \\
\addlinespace[2pt]
9 & \texttt{<think>} I could have misinterpreted important facts or overlooked necessary details. & 24 & \texttt{<think>} It's conceivable that I've neglected critical information or erred. \\
\addlinespace[2pt]
10 & \texttt{<think>} Aware that my reasoning might be flawed or lacking crucial points. & 25 & \texttt{<think>} Wondering if I've mistakenly dismissed something important or misunderstood it. \\
\addlinespace[2pt]
11 & \texttt{<think>} I need to reconsider—I might've skipped vital information or erred. & 26 & \texttt{<think>} Maybe my previous reasoning has blind spots or unnoticed errors. \\
\addlinespace[2pt]
12 & \texttt{<think>} There's a chance my previous thinking has unnoticed mistakes or omissions. & 27 & \texttt{<think>} I'm doubting if crucial points were missed or mistakes made earlier. \\
\addlinespace[2pt]
13 & \texttt{<think>} I feel there might be something critical I overlooked or misunderstood. & 28 & \texttt{<think>} Feeling uncertain—perhaps critical details slipped past or were misunderstood. \\
\addlinespace[2pt]
14 & \texttt{<think>} Perhaps my earlier reasoning has hidden mistakes or missing information. & 29 & \texttt{<think>} Recognizing possible gaps or missteps I didn't previously notice. \\
\bottomrule
\end{tabular}
\end{table*}

\section{Limitations, Discussion, and Future Work}
\label{app:sec:future_work}

\paragraph{Limited Exploration Strategy}
Our current exploration mechanism relies on a relatively simple strategy—injecting prompts from a pre-constructed prompt pool to guide the model toward alternative reasoning paths. While effective, this approach may fall short of the full potential of more sophisticated exploration techniques. From the prompt perspective, online generation of exploration prompts conditioned on the model’s current reasoning state may offer greater adaptivity and contextual relevance than our static prompt set. From the policy perspective, incorporating more structured search procedures, such as backtracking trees or trajectory-level search algorithms, could enable more systematic exploration across the reasoning space. Moreover, our method emphasizes local trajectory perturbation via prompt insertion, rather than global restructuring of the reasoning path. Despite these limitations, our results demonstrate that end-to-end optimization under an exploratory policy is both feasible and beneficial, laying the groundwork for future work on more principled and expressive exploration strategies.

\paragraph{Computational Overhead and Adaptive Sampling Limitations}
The mixed sampling strategy inherently introduces computational overhead compared to standard policy optimization approaches. Our resampling mechanism requires a two-stage process: first performing normal sampling to assess question difficulty through initial trajectory evaluation, then conducting exploratory sampling based on the observed failure rates. This sequential approach increases computational complexity as it necessitates generating $(1-\alpha)G$ additional exploratory trajectories from the probe policy $\pi_\varepsilon$, resulting in approximately 12\% more trajectory sampling in our experiments. While this overhead is substantially more efficient than uniform oversampling approaches (which require 20\% additional trajectories for minimal gains), the computational cost scales linearly with the resampling parameter $p$ and the exploration ratio $\alpha$. A more efficient approach would involve predicting question difficulty a priori and automatically adjusting sampling quantities accordingly, eliminating the need for the initial sampling phase. However, developing reliable difficulty prediction mechanisms remains an open challenge. Furthermore, the policy realignment mechanism requires computing importance sampling ratios for each token, adding non-negligible computational complexity during training.

\paragraph{Lack of Validation in Broader Agentic Tasks}
While REX-RAG demonstrates consistent improvements across seven open-domain question answering datasets, its effectiveness has only been validated within the RAG (retrieval-augmented generation) framework. Our method specifically targets reasoning-intensive QA tasks where external information retrieval and multi-turn reasoning are tightly coupled. As such, it remains unclear whether the proposed exploration and policy realignment strategies generalize to broader agentic scenarios—such as tool use, web navigation, or embodied planning—where action spaces, environmental feedback, and task dynamics differ substantially. Extending our framework to these settings would require adapting both the structured interaction protocol and the rollout mechanism to accommodate more complex state-action transitions. Future work may explore the applicability of REX-RAG's core ideas beyond QA, investigating how exploration with distribution correction can benefit general-purpose decision-making agents.

\section{Structured Search Interaction Protocol}
\label{app:sec:interaction}

The structured search interaction protocol employed in REX-RAG follows the framework established by Search-R1~\cite{Search-r1}, which defines a systematic approach for integrating reasoning and retrieval operations through specialized tokens and prompt templates. The structured interaction protocol relies on four primary special tokens that delineate different phases of the reasoning and retrieval process:

\subsection{Special Tokens}
\label{app:sec:special_tokens}

\textbf{\texttt{<think>} and \texttt{</think>}} encapsulate the model's internal reasoning process, allowing it to engage in chain-of-thought reasoning without external interference. Within these tags, the model can perform logical deduction, analyze given information, identify knowledge gaps, and plan subsequent actions. This internal reasoning phase is crucial for determining when external retrieval is necessary and formulating appropriate search queries.

\textbf{\texttt{<search>} and \texttt{</search>}} trigger external information retrieval operations. When the model generates these tokens, the content within them is interpreted as a search query that is executed against the external knowledge base. This mechanism allows for dynamic knowledge acquisition during the reasoning process.

\textbf{\texttt{<information>} and \texttt{</information>}} contain the retrieved external knowledge that is returned by the search engine in response to search queries. This mechanism allows for dynamic knowledge acquisition during the reasoning process.

\textbf{\texttt{<information>} and \texttt{</information>}} contain the retrieved external knowledge that is returned by the search engine in response to search queries. The content within these tags represents the top search results that are automatically inserted into the model's context after a search operation. This information serves as additional context that the model can analyze and incorporate into itse content within these tags represents the top search results that are automatically inserted into the model's context after a search operation. This information serves as additional context that the model can analyze and incorporate into its reasoning process.

These special tokens serve multiple purposes: they provide clear demarcation between different operational phases, enable selective training on specific components of the reasoning process, and facilitate systematic evaluation of reasoning quality versus retrieval effectiveness.

\textbf{\texttt{<answer>} and \texttt{</answer>}} mark the final response generation phase, where the model synthesizes information from both its internal reasoning and retrieved external knowledge to produce a comprehensive answer. The content within these tags represents the model's final output, incorporating insights gained througning and retrieval process.

These special tokens serve multiple purposes: they provide clear demarcation between different operational phases, enable selective training on specific components of the reasoning process, and facilitate systematic evaluation of reasoning quality versus retrieval effectiveness.

\subsection{System Prompt}
\label{app:sec:system_prompt}

The system prompt template structure orchestrates the interaction between reasoning and retrieval components through a carefully designed format that guides the model's behavior throughout the question-answering process. The template follows a think-search-answer paradigm that promotes systematic problem-solving. The entire prompt template is demonstrated in Fig.~\ref{app:fig:prompt_template}.

\section{Prompt Templates and Examples}
\label{app:sec:prompt_template}

The revision prompts are formulated to express uncertainty and encourage critical self-evaluation without being overly prescriptive. Each prompt is designed to maintain the model's natural reasoning flow while introducing a reflective pause that can lead to error correction and improved reasoning quality.

The Table~\ref{app:tab:revision_prompts} presents all 30 revision prompts used in our implementation. These prompts are randomly selected during training to provide diverse expressions of self-doubt and reflection.

\end{document}